%% file: main.tex
\pdfoutput=1

\documentclass[11pt]{article}

\usepackage{ACL2023}

\usepackage{times}
\usepackage{latexsym}

\usepackage[T1]{fontenc}

\usepackage[utf8]{inputenc}

\usepackage{microtype}

\usepackage{inconsolata}

\usepackage{colortbl}
\usepackage[ruled,vlined]{algorithm2e}
\usepackage{graphicx}
\usepackage{wrapfig}
\usepackage{amsmath}
\usepackage{amssymb}

\usepackage{bm,bbm,upgreek}
\usepackage{amsthm}
\usepackage{pifont}
\usepackage{xcolor}
\usepackage{caption}
\usepackage{subcaption}
\usepackage{floatrow}
\usepackage{url}
\usepackage{booktabs}
\usepackage{multirow}
\usepackage{wrapfig}
\usepackage{floatflt}
\usepackage{lipsum}
\usepackage{makecell}
\usepackage{mathtools}
\usepackage{dblfloatfix}
\usepackage{enumitem}
\usepackage{fdsymbol}
\newcommand{\cmark}{\textcolor{green!80!black}{\ding{51}}}
\newcommand{\xmark}{\textcolor{red}{\ding{55}}}
\newcommand{\ourdataname}{CausalDialogue}

\title{\textcolor{teal}{Causal}\textcolor{orange}{Dialogue}: Modeling Utterance-level Causality in Conversations}

\author{
  Yi-Lin Tuan$^\clubsuit$\quad Alon Albalak$^\clubsuit$\quad  Wenda Xu$^\clubsuit$\quad  Michael Saxon$^\clubsuit$\quad Connor Pryor$^\vardiamondsuit$\\
  {\bf Lise Getoor$^\vardiamondsuit$\quad William Yang Wang$^\clubsuit$}\\
  $^\clubsuit$ University of California, Santa Barbara, $^\vardiamondsuit$ University of California, Santa Cruz\\
  \texttt{\{ytuan, alon\_albalak, wendaxu, saxon, william\}@cs.ucsb.edu}\\
  \texttt{\{cfpryor, getoor\}@ucsc.edu}\\
}

\begin{document}

\maketitle

\begin{abstract}
    \input{sections/00abstract}
\end{abstract}

\input{sections/01introduction}

\input{sections/02related_work}

\input{sections/03data_collection}

\input{sections/04-1task_definition}

\input{sections/04-2models}

\input{sections/05experiments}

\input{sections/06analysis}

\input{sections/07conclusion}

\input{sections/08limitation.tex}

\section*{Acknowledgement}
This work was supported in part by the National Science Foundation under \#2048122 and an unrestricted gift award from Google Deepmind. The views and conclusions contained in this document are those of the authors and should not be interpreted as representing the sponsors.

\bibliography{main}
\bibliographystyle{acl_natbib}

\clearpage
\appendix
\input{sections/09appendix.tex}

\end{document}

%% file: sections/00abstract.tex
Despite their widespread adoption, neural conversation models have yet to exhibit natural chat capabilities with humans.
In this research, we examine user utterances as \textit{causes} and generated responses as \textit{effects}, recognizing that changes in a cause should produce a different effect.
To further explore this concept, we have compiled and expanded upon a new dataset called \textbf{\ourdataname{}} through crowd-sourcing.
This dataset includes multiple cause-effect pairs within a directed acyclic graph (DAG) structure.
Our analysis reveals that traditional loss functions struggle to effectively incorporate the DAG structure, leading us to propose a causality-enhanced method called Exponential Maximum Average Treatment Effect (ExMATE) to enhance the impact of causality at the utterance level in training neural conversation models.
To evaluate the needs of considering causality in dialogue generation, we built a comprehensive benchmark on \ourdataname{} dataset using different models, inference, and training methods.
Through experiments, we find that a causality-inspired loss like ExMATE can improve the diversity and agility of conventional loss function and there is still room for improvement to reach human-level quality on this new dataset.
\footnote{Our code and dataset are available at \url{https://github.com/Pascalson/CausalDialogue}}

%% file: sections/01introduction.tex
\section{Introduction}

Over time, broadly-defined dialogue models have become increasingly prevalent in society and been integrated in a range of domains from speech assistants and customer service systems to entertainment products, such as video games, where the non-playable characters (NPCs) engage in conversation with players.
A core goal of training chatbots is enabling them to interact with humans naturally~\cite{vinyals2015neural,sordoni2015neural}.
This includes, but is not limited to: considering both the machine and addressee's personalities~\cite{li2016persona}, diversifying responses to be less generic (e.g., the same response ``I don't know.'' is often produced in a traditional setting for different dialogues)~\cite{li2016diversity}, grounding on external knowledge to be informative~\cite{ghazvininejad2018knowledge},
and tailoring responses specific to nuanced differences in conversation.

To the best of our knowledge, no recent studies have prioritized the ability to tailor responses for minor differences in conversations.
This problem is currently implicitly approached by training models with larger scale or cleaner conversation data~\cite{zhang2020dialogpt,roller2021recipes,thoppilan2022lamda} or involving human-in-the-loop~\cite{li2016dialogue, jaques2020human}.
However, the effectiveness of these methods is unclear, the online rewarding scheme can be expensive, and a suitable testbed for evaluating the solution to this problem has not yet been identified.

\begin{figure*}[t]
    \centering
    \includegraphics[width=.95\linewidth]{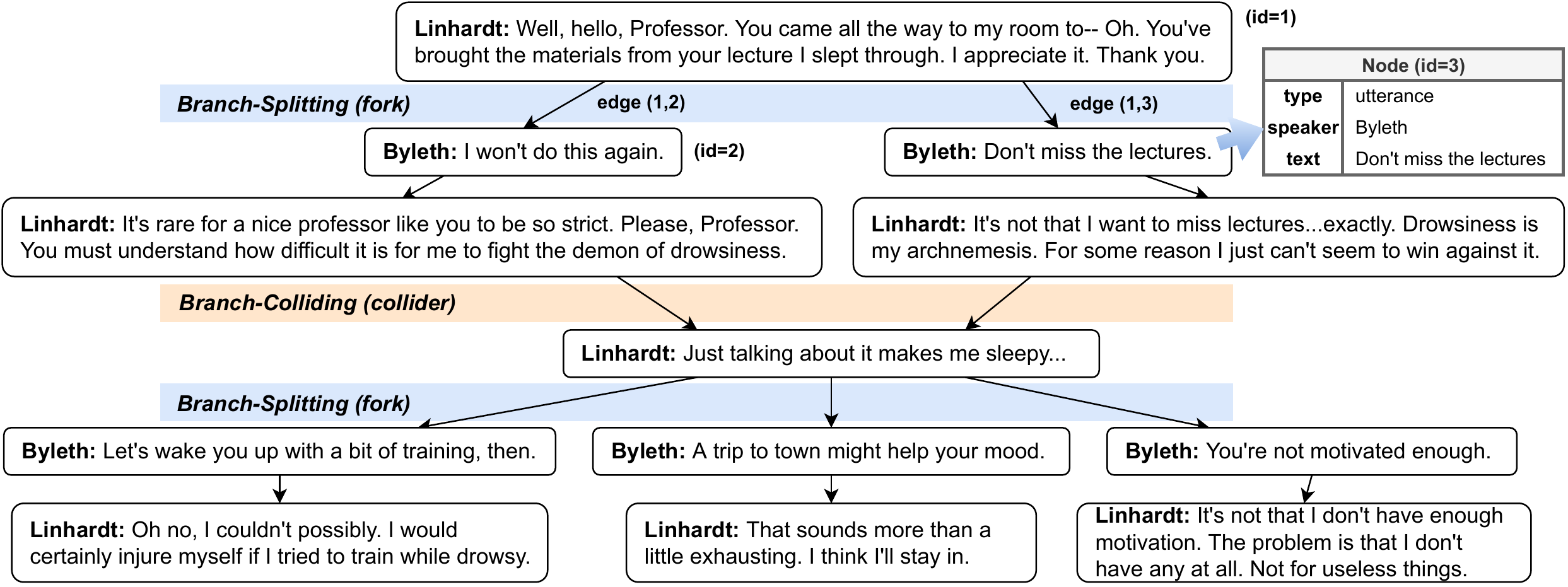}
    \caption{A dialogue DAG example in the new dataset \ourdataname{}. As the conversation progress, each utterance can be continued with multiple responses (branch-splitting; fork); meanwhile, the same root dialogue with different middle turns can be continued by the same response (branch-colliding; collider).}
    \label{fig:original-ddag-example}
\end{figure*}

To this end, we propose a benchmark to foster research in tailoring responses for nuanced differences in conversations by answering the question  ``{\it if all prior turns are the same, but the last turns in two conversations are semantically different, how should future turns differ?}''
We call this concept {\it Agility} and model it as the {\it utterance-level causes and effects} in dialogue response generation, where the causes are the slightly different prior turns and the effects are the resulting future turns.

We introduce {\bf \ourdataname{}}, a dataset seeded by expert-written dialogues containing branching dialogue paths, which we further expand in terms of scale and linguistic abundance with crowd-sourcing. 
Each conversation is represented as a directed acyclic graph (DAG) for ease of storage and causal analysis~\cite{pearl2009causality} as shown in Figure~\ref{fig:original-ddag-example}.
As conversations progress, each utterance can elicit multiple responses, resulting in a split of the conversation (branch-splitting). Alternatively, multiple conversations that share a common starting point may sometimes lead to the same response, even if the middle exchanges differ (branch-colliding).
Due to the DAG structure of \ourdataname{}, it is ideal for aiding research on response generation that requires abundant {\it IF}-bases, for instance, causal inference and offline reinforcement learning, which may improve the response generation quality for nuanced differences in conversation. 

\begin{table*}[t]\small
    \centering
    \begin{tabular}{c|cccccc}\toprule[1pt]
        & \bf \ourdataname{} & \bf TV Series & \bf MultiTalk & \bf DailyDialog & \bf PersonaChat & \bf LIGHT\\
        \midrule[0.5pt]
        Branches & \cmark (DAG) & \xmark & \cmark (Tree) & \xmark & \xmark & \xmark\\
        Profiles & \cmark & \cmark & \xmark & \xmark & \cmark & \cmark\\
        Situated & \cmark & \cmark & \cmark & \xmark & \xmark & \cmark\\
        Expert involved & \cmark & \cmark & \xmark & \cmark & \xmark & \xmark\\
        \bottomrule[1pt]
    \end{tabular}
    \caption{Compared to current widely used datasets, \ourdataname{} contains the utterance-level graph structure and meanwhile has the features of diverse speaker profiles, descriptive situations, and high quality scripts written by experts. The referring dialogue generation datasets are: TV series~\cite{tuan2019dykgchat}, MultiTalk~\cite{dou2021multitalk}, DailyDialog~\cite{li2017dailydialog}, PersonaChat~\cite{zhang2018personalizing}, LIGHT~\cite{urbanek2019learning}.}
    \label{tab:data_comparison}
\end{table*}

To provide a benchmark for future work on the \ourdataname{} dataset, we conduct experiments with various setups.
We include both decoder-only and encoder-decoder transformer models pretrained on either common or dialogue-specific corpora, various inference methods, conventional training losses, and a newly proposed loss, Exponential Maximum Average Treatment Effect (ExMATE), inspired by Average Treatment Effect~\cite{holland1986statistics,imai2008misunderstandings}, which is a method commonly used to approximate the causal effect of a treatment and its outcome.
In this benchmark, we show that existing methods are not sufficient in tackling the agility issue, and a simple causality-inspired loss demonstrates improvement.

Our key contributions are:

\setlength{\leftmargini}{10pt}
\setlist{nosep}

\begin{itemize}
\item A novel dataset, \ourdataname{}, including both expert-written scripts and crowd-sourced utterances with a DAG structure.
\item A new training loss, ExMATE, for considering the utterances as causes and effects in a dialogue, inspired by the average treatment effect in research on causal inference.
\item A benchmark with experiments showing that existing methods need improvement on the agility problem, and a causality-inspired method can be a promising direction to improve it.
\end{itemize}

%% file: sections/02related_work.tex
\section{Related Work}

\paragraph{Chit-Chat Dialogue Datasets.}
To boost the research of dialogue models, the community has collected dialogues based on scripts written by experts from movies~\cite{Danescu-Niculescu-Mizil+Lee:11a, banchs2012movie, lison2016opensubtitles2016}, TV shows~\cite{poria2019meld, tuan2019dykgchat, yu2020dialogue, rameshkumar2020storytelling}, and education purposes~\cite{li2017dailydialog,cui2020mutual}.
For abundant diversity and real-life scenarios, \citet{ritter2011data, wang2013dataset, lowe2015ubuntu, pasunuru2018game} collected datasets based on the publicly available data from social media and forums.
Additionally, previous work has explored the idea of collecting data through crowd-sourcing with added constraints to improve its quality or expand label types.
For example, \citet{zhang2018personalizing} constructed a dataset with workers imitating a given personal profile.
\citet{rashkin2019towards} built a dataset by explicitly asking workers to show their empathy during a conversation.
\citet{urbanek2019learning,narayan2019collaborative, ammanabrolu2021motivate} created datasets with the assistance of game structures, so the purpose of the dialogue is to complete a mission or collaborate with other agents.
Finally, recent work by \citet{dou2021multitalk} collected branches of dialogues for 120 self-written prompts to create dialogue trees.
Compared to previous studies, our dataset is a fusion of the scripts written by experts and responses created by crowd-sourcers with manual correction, granting it high quality, linguistic abundance, and extensive metadata.
Additionally, our dataset includes both branch-splitting and branch-colliding instances, which has led us to classify dialogues as directed acyclic graphs (DAGs) instead of just sequences or trees.

\paragraph{Dialogue Generation Training Objectives.}
To train a dialogue response generation model, methods have been developed from maximizing the likelihood between the hypothesis and the ground-truth~\cite{vinyals2015neural,serban2016building}, guiding responses to match a higher reward in reinforcement learning~\cite{li2016deep},
and allowing for extra latent variables to optimize divergence through variational autoencoder~\cite{zhao2017learning} or generative adversarial networks~\cite{li2017adversarial,tuan2019improving}.
Recent works have introduced the concept of causal inference~\cite{holland1986statistics,imai2008misunderstandings,pearl2009causality,cunningham2021causal} into generative adversarial network-based~\cite{zhu2020counterfactual} and multiple-stage inference based dialogue generation model~\cite{tuan2020knowledge}.
Utterance-level offline reinforcement learning has also been explored to optimize response generation~\cite{jaques2020human,verma-etal-2022-chai}.
However, they were studied by expanding available sequence data with imaginations.
Now by providing a chit-chat dialogue DAG structure that is enriched with multiple if-else cases, \ourdataname{} can be studied for causal inference and offline reinforcement learning on response generation.
We also propose a new method called ExMATE for better optimizing a response generation model on the DAG data structure.

%% file: sections/03data_collection.tex
\section{\ourdataname{} Dataset}
In this section, we introduce \textbf{\ourdataname{}}, a novel dataset that includes chit-chat conversations in a Conversational Directed Acyclic Graph (DAG) data structure.
This structure allows for the natural inclusion of various dialogue flows, such as forks (branch-splitting) and colliders (branch-colliding)~\cite{pearl2009causality}.
Our goal is to offer researchers a valuable resource for studying the complexities of human conversation and advancing the understanding of causal inference in dialogue.

To create \ourdataname{}, we sourced expert-written dialogues from a role-playing game (Section~\ref{sec:data_collection}) and expanded upon them with Amazon Mechanical Turk (MTurk)\footnote{\url{https://www.mturk.com}} and manual correction (Section~\ref{sec:data_expansion}).
By using our fused collection method, the dataset contains high-quality, engaging conversations with abundant linguistic usage that imitates daily life.

\subsection{Data Collection}
\label{sec:data_collection}

\ourdataname{} is derived from the English scripts of the popular role-playing game (RPG) \textit{Fire Emblem: Three Houses}, which we sourced from the fandom wikipedia\footnote{\url{https://fireemblem.fandom.com/}} under the GNU Free Documentation License(GFDL)\footnote{\url{https://fireemblem.fandom.com/wiki/Fire_Emblem_Wiki:Copyrights}}.
This RPG is well-known for its diverse, story-driven conversations, which mix the interactions of approximately 40 main characters.
In this game, players have the ability to shape the narrative by making choices that lead to different dialogue branches.

Table~\ref{tab:data-stats} lists the statistics of the two main types of the crawled data, which are already divided in the raw scripts.
We name the first conversation type \textsc{Ori.-2S}, which are mostly dialogues between two speakers, and generally include conversations about interpersonal relationships.
We name the second conversation type \textsc{Multi}, which are dialogues between two or more speakers, and usually describe the current status of the story line.
In the following sections, we will introduce the DAG structure to better describe the dataset, as well as how we obtained additional examples from crowd-sourcing to create the \textsc{Expansion} to these expert-written scripts.

\begin{table}[t]\small
    \centering
    \begin{tabular}{@{} l|rrr|r @{}}\toprule[1pt]
        \bf Data Partition & \textsc{\bf Ori.-2S} & \textsc{\bf Multi} & \textsc{\bf Expan.} & \bf Total \\\midrule[0.5pt]
        \# Dialogues$^\dagger$ & 794 & 1528 & 623 & 2322 \\
        \# Branches & 1633 & 1298 & 2378 & 4866\\
        \# Utterances & 33247 & 13858 & 15728 & 46109 \\
        \# Speakers & 41 & 47 & 39 & 51\\
        Avg. utts/dial. & 17.0 & 51.4 & 5.6 & 26.8\\
        Avg. words/utt. & 18.4 & 17.8 & 11.8 & 16.5\\
        Avg. utts/spk. & 801.6 & 268.8 & 402.8 & 878.4\\
        \bottomrule[1pt]
    \end{tabular}
    \caption{The statistics of \ourdataname{} dataset, where the columns \textsc{Ori.-2S} and \textsc{Multi} are the crawled and cleaned original scripts and the column \textsc{Expansion} is from crowd-sourcing. In total, there are 3457/741/715 dialogues for train/validation/test sets. $\dagger$ indicates that for \textsc{Expansion} set, 623 is the number of initial dialogues that are parts of the 794 \text{Ori.-2S} dialogues, so the total number of dialogues is 2322.}
    \label{tab:data-stats}
\end{table}

\paragraph{Dialogue DAGs.}

Conventional linear dialog data structures can be challenging to create when dealing with {\it forks} and {\it colliders}, as they can lead to ambiguity in the form of duplicated utterances and split responses.
To address this issue, we propose using a conversational DAG to maintain the fidelity of the dialog.
We convert each textual conversation into a DAG, as demonstrated in Figure~\ref{fig:original-ddag-example}.
Formally, each node is a dictionary containing the text type (utterance/scene information), text, speaker, and its own id in the dialogue.
A directed edge $(i,j)$ then indicates that a node with id $j$ is a possible response to the node with id $i$.
Saving dialogues as DAGs may introduce some complexity, but it also offers numerous benefits.
For example, it reduces the memory required to save each dialogue branch independently, enables a natural visualization of the multiple possible dialogues flows, and fosters the survey of causality on dialogue utterances.

\paragraph{Speaker Profiles.}
Prior work has shown the relationship between personality and language uses in conversations~\cite{mairesse2007using}.
To ensure consistent personality, as well as to diversify linguistic features across speakers, we leverage the speaker profiles during the data collection process.
The resulting \ourdataname{} dataset comprises 41 main speakers who have been thoughtfully crafted by the game's developers.
These speakers possess diverse backgrounds, perspectives, and interests, and their characteristics are both human-like and distinct.
These speaker profiles are simplified for collecting the \textsc{Expansion} partition to reduce workers' cognitive load, and a set of examples are provided in Appendix~\ref{apx:speaker_profile}.
Compared with the speaker profiles in \ourdataname{}, previous works have provided limited information (e.g. ``I have a dog.'')~\cite{zhang2018personalizing,urbanek2019learning}, or have a significantly smaller number of speakers~\cite{poria2019meld,tuan2019dykgchat}

\begin{figure}[t]
    \centering
    \includegraphics[width=.9\linewidth]{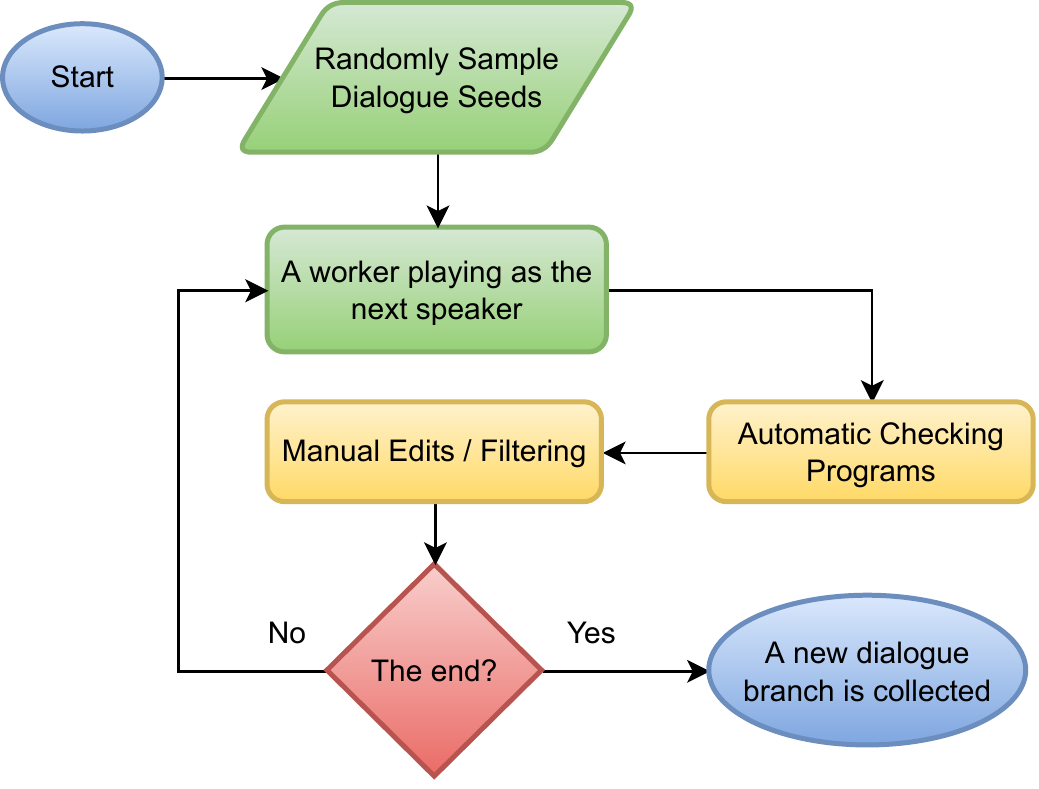}
    \caption{The flowchart of our strategy for data expansion with crowd-sourcing.}
    \label{fig:expansion-flowchart}
\end{figure}

\begin{figure*}[t]
    \centering
    \includegraphics[width=.98\linewidth]{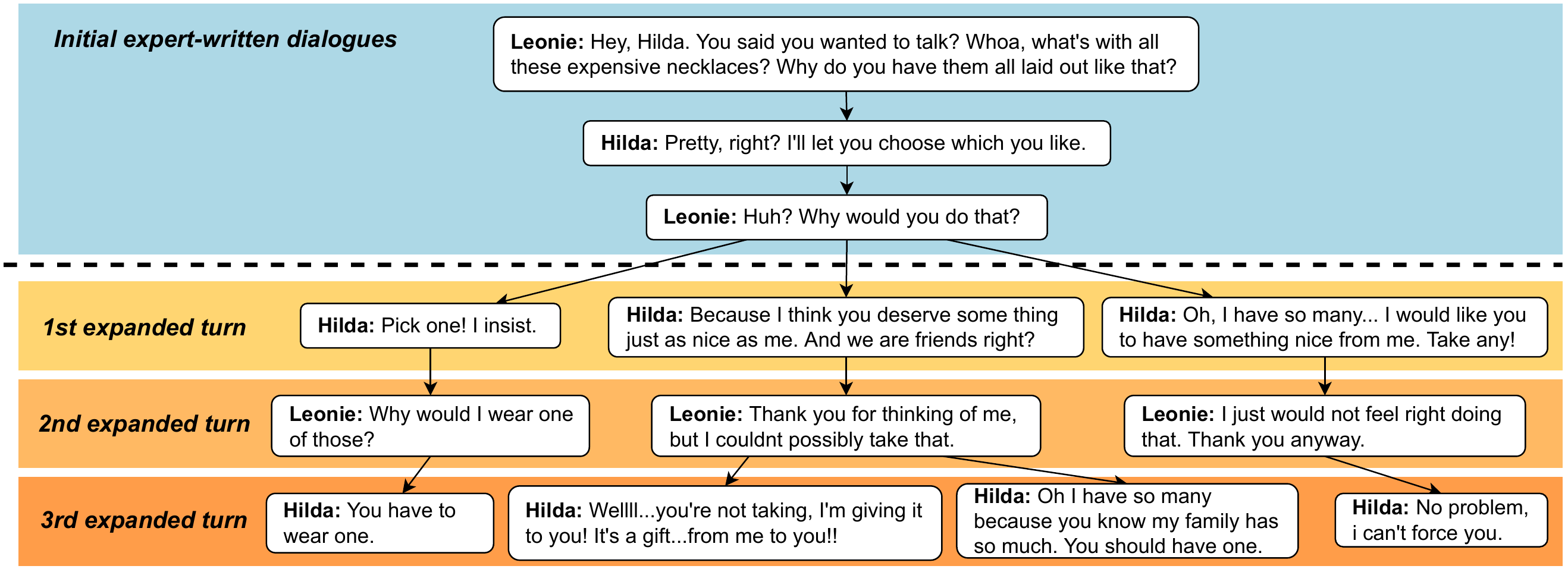}
    \caption{A dialogue example of the \textsc{Expansion} partition in \ourdataname{}.}
    \label{fig:expansion-ddag-example}
\end{figure*}

\subsection{Data Expansion}
\label{sec:data_expansion}

In order to increase the breadth and scope of our dataset, we propose utilizing a crowd-sourcing approach to add more diverse and current language as shown in Figure~\ref{fig:expansion-flowchart} (More details in Appendix~\ref{apx:data_expansion}). 

\paragraph{Initial Dialogue Selection.}

We first randomly select 1,200 partial dialogues from the \textsc{Ori.-2S} partition, which is of higher quality after our manual inspection.
This can result in more stable quality when crowd-sourcing responses.

\paragraph{Expansion Collection.}
Each initial dialogue along with the continuing speaker profile is presented to 3 workers on MTurk to write the next utterance.
A new branch of continued dialogue will then be presented to another 1-2 workers playing as another speaker to gather another round of responses.
We repeated this process three times and collect a total of about 13,000 written utterances.
Table~\ref{tab:data-stats} lists the detailed statistics of the expanded data in the column \textsc{Expansion}.
Note that the statistics of \textsc{Expansion} in Table~\ref{tab:data-stats} include the initial dialogues.
Figure~\ref{fig:expansion-ddag-example} shows an DAG representation of an expanded example.

\paragraph{Quality Control.}
We adopt three strategies to control for dialogue quality.
First, we asked the workers on MTurk to annotate if they regard a dialogue as already completed or having too specific of details to continue.
The purpose of the first stage of quality control is to identify conversations which cannot be continued, either because the conversation has already concluded or because the workers are lacking enough information about the world to continue the conversation.
Second, we used an off-the-shelf model\footnote{\url{https://github.com/unitaryai/detoxify}} to label potential ethical issues inside the collected utterances for reference in the next step.
Finally, we invited real players of the game and machine learning researchers to manually check all the utterances by their fluency, coherence, and ethics as well as referring to the labels from the previous two steps to ensure the final \textsc{Expansion} partition is of high quality.

%% file: sections/04-1task_definition.tex
\section{Task Definition}
\label{sec:task_def}
In this work we consider a conversation among two or more speakers.
At each time step $t$, a speaker $s_t$ takes their turn with an utterance $u_t$.
The goal, as in conventional response generation, is to train a model parameterized by $\theta$ that can predict a plausible next utterance given the speakers and utterances in prior turns as:
\begin{equation}
    u_{t+1} \sim P_\theta(\cdot| s_1 u_1, s_2 u_2, ..., s_t u_t, s_{t+1})\,.
\end{equation}

Distinct from prior conversation datasets, \ourdataname{} has multiple dialogue branches.
If we consider each branch as an independent conversation (flatten the branches), many conversations will have large overlaps and thus bias the dataset.
We consider this point and extract triples $(DH, x, y)$ from \ourdataname{}.
To simplify notations for following sections, we denote $s_t u_t$ as $x$, $s_{t+1} u_{t+1}$ as $y$ and $DH$ is the dialogue history $s_1 u_1, s_2 u_2,..., s_{t-1} u_{t-1}$.
The key idea is that for a $DH$, we will not extract duplicated pairs $(x,y)$, but $x$ or $y$ itself can be shared.

The \ourdataname{} response generation task is therefore defined as finding a possible turn-taking speaker and their response given the dialogue history $DH$ with an utterance cause $x$.
\begin{equation}\label{eq:task-def}
    y \sim P_\theta(\cdot | DH,x)\,.
\end{equation}
The sequences $x=x_1x_2...x_i...x_{|x|}$ and $y=y_1y_2...y_j...y_{|y|}$, where $x_i$ and $y_j$ are tokens, and $|x|$ and $|y|$ are the length of the sequences $x$ and $y$ respectively.

\subsection{Agility}
\label{sec:agility}

While the above task definition resembles the standard dialogue generation setting with the exception of speaker prediction and conversation overlaps, our primary interest lies in tailoring responses to minor differences in conversation history.
We refer to this concept as {\it Agility}, where a minor difference in conversations can be a shared $DH$ with different continuation $x$.

To quantify the idea of agility, we propose a new metric with the following idea:
If the predicted next utterance $y$ and the previous turn $x$ has causal-effect relationship (i.e., $x_1\rightarrow y_1$ and $x_2\rightarrow y_2$), we anticipate that it is less likely that $y_2$ is caused by $x_1$.
The newly proposed metric, named confidence causal-effect (CCE) is formally defined as:
\begin{equation}
\begin{split}
    CCE = & E_{(x,y)\in D, (x,y')\notin D, (x',y')\in D}\\
    & [PPL_\theta(y'|DH,x) - PPL_\theta(y|DH,x)]\,,
\end{split}
\end{equation}
where PPL refers to perplexity.
Note that CCE is not a metric that stands by itself and needs to refer to PPL at the same time.
That is, given a similar PPL score, a model with higher CCE score is better.
Additionally, it is important to acknowledge that the concept of agility has been indirectly incorporated into conventional dialogue generation models and evaluation metrics, but it has not been specifically examined in isolation.
Our newly introduced dataset and CCE metric can be seen as an initial step towards addressing this aspect.

%% file: sections/04-2models.tex
\section{Methods}
In this section, we describe how conventional generative models can be used and propose a simple yet effective approach to model causal effect.

\subsection{Maximize Likelihood Estimation}
An often used method to train a conditional sequence generation model is minimizing the negative log likelihood~\cite{vinyals2015neural,serban2016building}.
The loss function is as following:
\begin{equation}
\begin{split}
    & L_{MLE} = \\
    & \underset{(DH,x,y)\sim P_D}{E} \sum_{j=1}^{|y|} -\log P_\theta(y_j | DH,x,y_{1...j-1})\,,
\end{split}
\end{equation}
where $P_D$ represents the data distribution.
Since the duplication of dialogue history is already taken in to account in our task definition (Section~\ref{sec:task_def}), this MLE method can be seen as the recently proposed dialogue tree model~\cite{dou2021multitalk}.
However, this function only models a part of the cause-effect relationship between the condition and the output sequence.
This neglect may lead to a more vague predicted probability distribution of the output, thus generating less agile responses.

\subsection{Maximize Average Treatment Effect}

To explicitly model the causal effect in a conversation, we propose the Exponential Maximum Average Treatment Effect (ExMATE), taking into account the treatment effect in causal inference~\cite{pearl2009causality}.
The treatment effect, denoted by $\delta$, is defined as the difference between the outcome under treatment $I=1$, represented by $\mathcal{O}^{I=1}$, and the outcome under treatment $I=0$, represented by $\mathcal{O}^{I=0}$.
This measures the variation in outcomes when an event $I$ is present or absent.
A higher value of $\delta$ indicates that the event $I$ is more likely to be a true cause of the outcome. Conversely, a small value of $\delta$ suggests that the event $I$ is unlikely to be a cause of the outcome and may only be correlated.
We aim to utilize this characteristic in dialogue generation modeling to ensure that a preceding utterance can be considered the genuine cause of the predicted response.

\begin{figure}[t]
    \centering
    \includegraphics[width=0.5\linewidth]{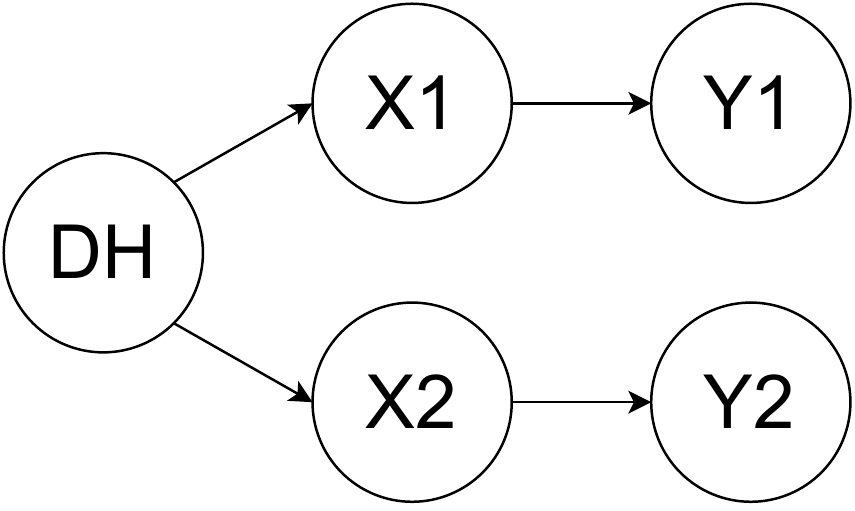}
    \caption{The graphical model of fork-like DAG considered in our proposed ExMATE loss.}
    \label{fig:fork-DAG}
\end{figure}

We consider the {\it fork-like} DAGs (as shown in Figure~\ref{fig:fork-DAG}) existing in a dataset such as Figure~\ref{fig:original-ddag-example} and Figure~\ref{fig:expansion-ddag-example}.
Without loss of generality, in a binary case, this type of DAG involves two triples that share the same $DH$ and can be simplified as having nodes $DH$, $X_1$, $X_2$, $Y_1$, and $Y_2$.
Here we use ($X_1$, $Y_1$) and ($X_2$, $Y_2$) to denote two possibilities of ($x$,$y$) after $DH$.
We take $I=1$ as choosing the branch $X_1$, and $I=0$ as choosing an alternative branch $X_2$.
Therefore, a traditional definition of the treatment effect $\delta_i=|\mathcal{O}_i^{I=1}-\mathcal{O}_i^{I=0}|$ for the $i$-th example in this type of DAG can be rewritten as:

\begin{equation}
    \delta_i \triangleq \underset{\substack{X_1\sim P_D(\cdot|DH_i),\\ X_2\sim P_D(\cdot|DH_i),\\X_1 \neq X_2}}{E} |\mathcal{O}_i^{X_1} - \mathcal{O}_i^{X_2}|\,,
\end{equation}
where $\mathcal{O}^{X_1}_i$ or $\mathcal{O}^{X_2}_i$ is the outcome of an oracle given $X_1$ or $X_2$ as the input.

Since the outcome of a dialogue model is hard to be mathematically described only by an input $X$,
we instead utilize the uncertainty of predicting the pair $(x,y)$ by a model $\theta$.
We abuse the notation $\mathcal{O}_i$ here and redefine it as,
\begin{equation}\label{eq:our_outcome_def}
    \mathcal{O}_{i,Y_1}^{X_1} \triangleq P_\theta(Y_1|DH,X_1)\,.
\end{equation}

\begin{table*}[t]\small
    \centering
    \begin{tabular}{lllcccccc
    >{\columncolor[gray]{.8}[0pt]}c c}\toprule[1pt]
    
        \multirow{2}{*}{\bf Model} & \multirow{2}{*}{\bf Loss} & \multirow{2}{*}{\bf Inference} & \multicolumn{4}{c}{Fluency} & \multicolumn{2}{c}{Diversity} & \multicolumn{1}{>{\cellcolor[gray]{.8}}c}{Agility} & \multicolumn{1}{c}{Identity}\\
        & & & PPL ($\downarrow$) & BLEU1 ($\uparrow$) & 2 ($\uparrow$) & 4 ($\uparrow$) & Dist1 & Dist2 & CCE ($\uparrow$) & Acc ($\uparrow$)\\\midrule[0.5pt]
        \multicolumn{3}{l}{Human Written Responses} & 1.2 & 48.9 & 34.0 & 25.9 & 1.70 & 11.1 & Inf & 100.0 \\\midrule[0.5pt]
        DG & MLE & Greedy Search & 18.9 & 11.2 & 4.47 & 0.84 & 0.73 & 3.42 & 2.33 & 32.51 \\
        DG & MLE & Softmax (T=0.5) & 18.9 & 17.0 & 6.43 & 1.17 & 1.12 & 9.09 & 2.33 & 30.97 \\
        DG & MLE & TopK (K=10) & 18.9 & 15.7 & 5.34 & 0.81 & 1.37 & 13.57 & 2.33 & 27.65 \\
        DG & ExMATE & Greedy Search & 19.0 & 10.7 & 4.26 & 1.05 & 0.79 & 3.65 & 2.68 & 32.18 \\
        DG & ExMATE & Softmax (T=0.5) & 19.0 & 15.5 & 5.70 & 1.06 & 1.25 & 9.71 & 2.68 & 31.18 \\
        DG & ExMATE & TopK (K=10) & 19.0 & 13.5 & 4.47 & 0.67 & 1.52 & 14.44 & 2.68 & 28.16 \\
        \midrule[0.5pt]
        T5 & MLE & Greedy Search & 15.4 & 5.80 & 2.52 & 0.58 & 1.11 & 4.37 & 1.39 & 75.64 \\
        T5 & MLE & Softmax (T=0.5) & 15.4 & 12.7 & 5.06 & 0.97 & 1.77 & 10.91 & 1.39 & 74.66 \\
        T5 & MLE & TopK (K=10) & 15.4 & 14.1 & 5.09 & 0.82 & 2.07 & 15.49 & 1.39 & 72.79 \\
        T5 & ExMATE & Greedy Search & 15.4 & 5.66 & 2.46 & 0.55 & 1.10 & 4.06 & 1.50 & 75.76 \\
        T5 & ExMATE & Softmax (T=0.5) & 15.4 & 12.6 & 5.02 & 1.00 & 1.72 & 10.73 & 1.50 & 74.80 \\
        T5 & ExMATE & TopK (K=10) & 15.4 & 14.1 & 5.06 & 0.80 & 2.06 & 15.67 & 1.50 & 72.83 \\
        \bottomrule[1pt]
    \end{tabular}
    \caption{The test results on \ourdataname{} of different fine-tuned backbone models (DialoGPT (DG) and T5), inference methods (Greedy Search, Softmax, TopK), and loss functions (MLE and ExMATE). Using ExMATE loss enhances the agility aspect of dialogue generation models without compromising their fluency ratings.}
    \label{tab:results-test}
\end{table*}

After formulating a dialogue generation problem as utterance-level causal analysis as above, we apply the Average Treatment Effects (ATE)~\cite{holland1986statistics} to conversational DAGs, which is defined as 
\begin{equation}\label{eq:our_ATE_def}
\begin{split}
        & ATE \triangleq E_i [\delta_i] = E_i [\delta_{i,Y_1} + \delta_{i,Y_2}]\\
        & = E_i [\mathcal{O}_{i, Y_1}^{X_1} - \mathcal{O}_{i, Y_1}^{X_2} + \mathcal{O}_{i, Y_2}^{X_2} - \mathcal{O}_{i, Y_2}^{X_1}]\,.
\end{split}
\end{equation}

Recall that our goal is to strengthen the cause-effect relationship of each pair, ($X_1$,$Y_1$) and ($X_2$,$Y_2$) in the binary case.
This can be taken as maximizing the defined ATE in Equation~\ref{eq:our_ATE_def} with respect to the model parameters $\theta$.

Therefore, we substitute the $\mathcal{O}_{i,Y}^X$ term in Equation~\ref{eq:our_ATE_def} with its definition stated in Equation~\ref{eq:our_outcome_def} and derive:
{\small
\begin{equation}
\begin{split}
    \arg\max_\theta & ATE = \\ 
    \arg\max_\theta & \underset{(X_i,Y_i)\sim P_D(\cdot|DH)}{E} P_\theta(Y_i|DH,X_i)\\
    & - \underset{\substack{X_i\sim P_D(\cdot|DH),Y_j\sim P_D(\cdot|DH)\\(DH,X_i,Y_j) \notin D}}{E} P_\theta(Y_j|DH,X_i)\,.
\end{split}
\end{equation}
}

To stabilize the training, we modify it with logarithmic and exponential terms and call it the ExMATE loss function. Formally, it is written as:

{\small
\begin{equation}
\begin{split}
    & L_{ExMATE} = \\
    & \underset{\substack{(DH,x,y)\sim P_D,\\x_c\sim P_D(\cdot|DH),\\(DH,x_c,y)\notin D}}{E} - \log P_\theta(y|DH,x) + \exp ( \log P_\theta(y|DH,x_c))\,.
\end{split}
\end{equation}
}
The intuition for this change is that without $\exp(\cdot)$, the gradient of the second term will dominate the loss function, since $\log(u)$ has much larger gradient for $u$ close to $0$ than $u$ close to $1$ and an $\exp(\cdot)$ term can linearize it.

Overall, the idea of ExMATE is to maximize the response generation model's causal effects given a specific $X_i$ (or $(DH,x)$) as the current cause.
At the end, we found that this ATE-inspired approach turns out to be a combination of MLE and a subtraction of specific negative samples.
This formulation shares a similar concept with negative sampling and contrastive learning~\cite{goldberg2014word2vec,chen2020simple}, but has different example selection scheme and is not applied on the embedding space.
With this method, we are interested in the research question: {\it Will a model trained on the \ourdataname{} dataset be affected when using a causality-inspired loss?}

%% file: sections/05experiments.tex
\section{Experiments}

We provide a preliminary benchmark for \ourdataname{} with often used methods and a naive causality-inspired loss.
We fine-tuned two types of pretrained language models based on transformers~\cite{vaswani2017attention}: decoder-only architecture, DialoGPT~\cite{zhang2020dialogpt} and encoder-decoder architecture, T5~\cite{raffel2020exploring}, by the conventional MLE loss and the proposed ExMATE loss, and inferred by various sampling methods.
We evaluate three aspects of the generated responses: Fluency (perplexity (PPL) and BLEU~\cite{papineni2002bleu}), Diversity (Distinct n-grams Dist1 and Dist2~\cite{li2016diversity}), and our proposed Agility (CCE) in Section~\ref{sec:agility}.
Furthermore, we use accuracy to evaluate if the speaker for a given turn is correctly predicted as the one in the human written responses (Identity Acc).
More details are in Appendix~\ref{apx:metrics} and \ref{apx:exp_setup}.

\begin{table*}[t]\small
    \centering
    \begin{tabular}{@{} l|l|p{0.4\linewidth}|p{0.4\linewidth} @{}}\toprule[1pt]
        \multicolumn{2}{l|}{\makecell[l]{\bf Dialogue\\\bf History ($DH$)}}
        & \multicolumn{2}{p{0.7\linewidth}}{\makecell[l]{Lysithea: Oh, hey. It's you. Going for a walk again today? \\ Ignatz: No, I'm on cooking duty today, and I have to head into town for some groceries.}}\\
        
        \midrule[0.5pt]
        
        \rowcolor[gray]{.8}
        \multicolumn{2}{l|}{\bf Cont. Conv ($x$)}
        & {\bf case1} Lysithea: That sounds like quite a task!
        & {\bf case2} Lysithea: Would you like some company?\\\midrule[0.5pt]
        
        \parbox[t]{3mm}{\multirow{4}{*}{\rotatebox[origin=c]{90}{\makecell{response($y$)}}}}
        &
        \bf MLE+S & Ignatz: You're right. & Ignatz: Yes, I'd love to.\\
        & \bf MLE+T & Ignatz: I'll go for a run. & Ignatz: No, I'm not.\\
        & \bf ExMATE+S & Ignatz: I'm not sure I can handle it. & Ignatz: Yes, I'm a bit tired of the whole thing. \\
        & \bf ExMATE+T & Ignatz: Well, I have to find a new name for myself. & Ignatz: No! It's just me and I.\\

        \midrule[0.5pt]
        
        \rowcolor[gray]{.8}
        \multicolumn{2}{l|}{\bf Cont. Conv ($x$)}
        & {\bf case3} Lysithea: Allow me to join you so I can help.
        & {\bf case4} Lysithea: All on your own? Hm, I'd better go with you. I'd be worried if you went by yourself.\\\midrule[0.5pt]
        
        \parbox[t]{3mm}{\multirow{4}{*}{\rotatebox[origin=c]{90}{\makecell{response($y$)}}}}
        &
        \bf MLE+S & Ignatz: Oh, I'm on cooking duty. & Ignatz: I'm just a bit worried about you.\\
        & \bf MLE+T & Ignatz: I'm going to help cook. & Ignatz: Yeah, no, not really! I can't help you, but I'm just going to help you out.\\
        & \bf ExMATE+S & Ignatz: I'm not working, but I'll help you. & Ignatz:  I've got no choice. I'll just have to go out alone.\\
        & \bf ExMATE+T & Ignatz: Oh. I'm sorry, I couldn't be there for you. & Ignatz: Is it okay?\\

        \bottomrule[1pt]
    \end{tabular}
    \caption{Generated responses by our trained models, T5 models trained by MLE or ExMATE inferred by Softmax(S) or TopK(T) sampling methods, given a shared dialogue history but different branches. Using ExMATE loss generally produces more diverse and agile responses.}
    \label{tab:gen-examples}
\end{table*}

\subsection{Results}
The test results of human written responses and models trained and inferred by different setups are listed in Table~\ref{tab:results-test}.
We provide details of how we evaluate human written responses in Appendix~\ref{apx:eval_human_responses}.

{\bf [Backbone Models]} We observe that our trained T5 model is generally better than DialoGPT model, as evidenced by the significant difference in PPL and Identity Acc between them.
{\bf [Inference Methods]} We observe that Softmax and TopK can achieve better results than greedy search in this dataset, as evidenced by their BLEU and Distinct-N scores.
The reason is similar to the conventional generic response problem in open-domain dialogue generation~\cite{li2016diversity,tuan2019improving}, since in a DAG, a ($DH$, $x$) pair have multiple $y$ as references, causing even an ideal probability distribution to have high entropy.
{\bf [Loss Functions]} We find that ExMATE improves MLE with better diversity, agility, and identity accuracy, while maintaining similar fluency scores.
This meets our expectation that ExMATE should not deteriorate the MLE's ability in training a model while maximizing the potential causal effect in response prediction.
This result empirically shows that the causal effect can help to increase diversity and predict the turn-taking speaker as well.
Finally, compared to the evaluation results of human written responses (a hard-to-reach upper bound), current methods still need improvement, except for diversity scores.

\subsection{Human Evaluation}
We randomly sample 100 dialogues, present each example to three workers on MTurk and ask them score the three dimensions, agility, coherence, and informativeness, scaling from 1 to 5.
The evaluation form is provided in Appendix~\ref{apx:human_eval_form}.
For each example, we present one shared dialogue history with two branches and the corresponding machine generated responses or a human written response.
We randomly mix the human written ones to validate if the human evaluation is reliable to an extent, by anticipating human written ones will get higher scores.
We list the average ratings in Table~\ref{tab:human-eval-results}.
The model trained with ExMATE achieves a similar informativeness level as human written ones, and gets a higher agility rating, which is its main goal.
However, ExMATE can compromises coherence due to the subtraction of a counter example, which is a natural sentence, in its objective function.
The human evaluation demonstrates the challenge of models to meet human-level quality in \ourdataname{} featured by conversational DAGs, a portion of the diversed types of flows in the real world.

\begin{table}[t]\small
    \centering
    \begin{tabular}{l|cc >{\columncolor[gray]{0.8}[0pt]}c}\toprule[1pt]
        \bf Model & Coherence & Informativeness & Agility \\\midrule[0.5pt]
        Human & 3.78 & 3.72 & 3.49\\
        MLE & 3.63 & 3.60 & 3.36\\
        ExMATE & 3.59 & 3.74 & 3.40\\
        \bottomrule[1pt]
    \end{tabular}
    \caption{The human evaluation results (scale 1-5, the higher the better) of models trained on \ourdataname{} (MLE, ExMATE), and human written responses (Human) for reference.}
    \label{tab:human-eval-results}
\end{table}

%% file: sections/06analysis.tex
\subsection{Qualitative Analyses and Discussion}

Table~\ref{tab:gen-examples} shows an example of a shared dialogue history, four different continuations (case1-4), and responses generated by the same backbone model, T5, trained with different objectives and inferred with different sampling methods. 
We observe that responses produced by MLE+T (TopK), ExMATE+S (Softmax), ExMATE+T are generally coherent to the conversation, while ExMATE often produces more diverse and agile responses to different continuation cases (different $x$).
It is notable that other than the improvements, we find that all the models have three types of issues: mode collapse, semantic repetition, and identity misplacement.
{\bf [Mode Collapse]}
The problem is often-seen when inferring a model by greedy search, specifically, the predicted responses often repeat the same phrase such as ``I'm not sure''.
While tacking the issue by adopting inference sampling, we conjecture the reason is that in a DAG, using a typical loss function can learn a probability distribution with higher entropy.
This also demonstrates the need of a new loss function for training on a conversational DAG dataset.
{\bf [Semantic Repetition]} An example is the MLE+T response in Table~\ref{tab:gen-examples} case 4, where ``can't help you '' and ``help you out'' have semantic overlaps.
This issue can possibly be mitigated by repetition reduction techniques, such as unlikelihood training~\cite{welleck2019neural} in future work.
{\bf [Identity Misplacement]} The problem happens when a model is confused about its position in a dialogue. For instance, the MLE+T response in Table~\ref{tab:gen-examples} case 3 is more like an utterance of speaker Lysithea instead of Ignatz.
This issue might be soothed by existing persona consistent techniques~\cite{li2016persona,mazare2018training,su2019personalized} for building a overall good chatbot, while in this work, we focus on proposing a new dataset to benchmarking on the agility issue.

%% file: sections/07conclusion.tex
\section{Conclusion}
In this paper, we presented a new dataset, \ourdataname{}, with novel conversational DAG structure.
With experiments on various model setups with a newly proposed loss, ExMATE, we demonstrate that there is room for improvement to reach human-level quality, even though ExMATE improves the diversity, informativeness, and agility.
This dataset serves as a testbed for future research that needs abundant conversation cases, like causal inference and offline reinforcement learning.
Moreover, with the naturally paired metadata, future work can use this dataset for other tasks, such as speaker prediction in multi-speaker scenarioes and personalized dialogue generation.

%% file: sections/08limitation.tex
\section*{Limitations}
The introduced dataset has a moderate scale, as it is currently designed for fine-tuning instead of large model pretraining.
Our proposed collection scheme can be futher applied to enlarge the dataset.
Moreover, as we focus on English, the data source has multiple language versions written by experts. Hence, extending CausalDialogue to multilingual is straightforward.
With reward labeling, the dataset can be more intuitively used for offline RL.
Meanwhile, the dataset includes personality descriptions that can be used for personalized dialogue generation, even though is not the focus in this paper.
Finally, training a generative model on dialogue domain can require various computational costs, depending on the aspects such as lengths of input and output texts and number of model parameters, as well as special designs to prevent misuses.

\section*{Ethics Consideration}
The dataset is based on RPG game in fantasy world with diverse scenarios, including wars.
To match the story background, a model trained on this dataset might produce war-related words.
We manually looked into each example to meanwhile keep each speaker's personality and remove utterances that could potentially cause negative impact, such as violence, bias, and offensive words.

For the data annotation part and human evaluation part, we utilized the Amazon Mechanical Turk platform and required workers to have a HIT Approval Rate of greater than
95\% and be located in CA or the US.
We pay the annotators over 16 US dollars per hour on average, which is above the highest state minimum wage.
Given our setting, the workers understood the scenarios and agreed that their annotations will be used for research.
The data annotation part of the project is classified as exempt by Human Subject Committee via IRB protocols.

%% file: sections/09appendix.tex
\section{Appendix}

\begin{table*}[t]\small
    \centering
    \begin{tabular}{@{} l|p{6cm}|p{7cm} @{}}\toprule[1pt]
        \bf Speaker & \bf Profile Excerpt & \bf Example Utterances \\\midrule[0.5pt]
        \multirow{2}{*}{Byleth} & Byleth has a very subdued personality and rarely expresses emotion. & - It's all right. // - Not really. // - I'm sorry.\\\midrule[0.5pt]
        \multirow{3}{*}{Edelgard} & Edelgard holds herself with a dignified air, but full of melancholy and solemn wistfulness. & - That's exactly right. There will no longer be lords who inherently rule over a particular territory. // - Perhaps not. Still, here you are. Maybe I can trust you with this... \\\midrule[0.5pt]
        \multirow{3}{*}{Claude} & Claude is described as easygoing on the surface, but has a side that forces others to keep their guard around him. & - Huh? Are you actually reading? I thought you hated studying. // - Was that story really worth bawling your eyes out over?\\
        \bottomrule[1pt]
    \end{tabular}
    \caption{In \ourdataname{} dataset, some speakers profiles excerpts and their example utterances in conversations.}
    \label{tab:speaker_profile_and_examples}
\end{table*}

\subsection{Speaker Profiles}
\label{apx:speaker_profile}

Table~\ref{tab:speaker_profile_and_examples} provides a few examples of the speakers' profiles and utterances.

\subsection{Data Expansion Details}
\label{apx:data_expansion}

\paragraph{Initial Dialogue Selection.}

We first randomly select $m$ dialogues with replacement from the \textsc{Ori.-2S} partition, which is of higher quality after our manual inspection.
This can result in more stable quality when doing crowd-sourcing.
For each sampled dialogue, we randomly select a start time stamp $t$ from $Poisson(\lambda=1)$.
Next, we adjust the sampled time stamp $t$ to make sure it lies in an appropriate point to continue the dialogue by $t^* = \max(\min(t+2, L),2)$, where $L$ is the maximum time stamp of this dialogue.
For each time stamp, if the original dialogue has multiple possible nodes, we select one randomly from a uniform distribution.
This process results in $m$ initial dialogues $D_0$ with various lengths (at least two utterances) for expansion.

\paragraph{Expansion Collection.}
Each initial dialogue $D_0$ along with the continuing speaker profile is presented to $n$ workers on MTurk to write the next utterance.
The new continued dialogues $D_1$ will then be presented to another 1-2 workers, decided by $p$\%, playing as another speaker to gather another round of responses.
This results in about $mn((1+p)^T - 1)/p$ new utterances for data expansion, where $T$ is the number of iterations.
Our expansion data is set with $m=1200$, $n=3$, $p=0.2$, and $T=3$. This setting results in about 13,000 written utterances.

\subsection{Human Annotations}

\begin{figure*}[t]
    \centering
    \includegraphics[width=.95\linewidth]{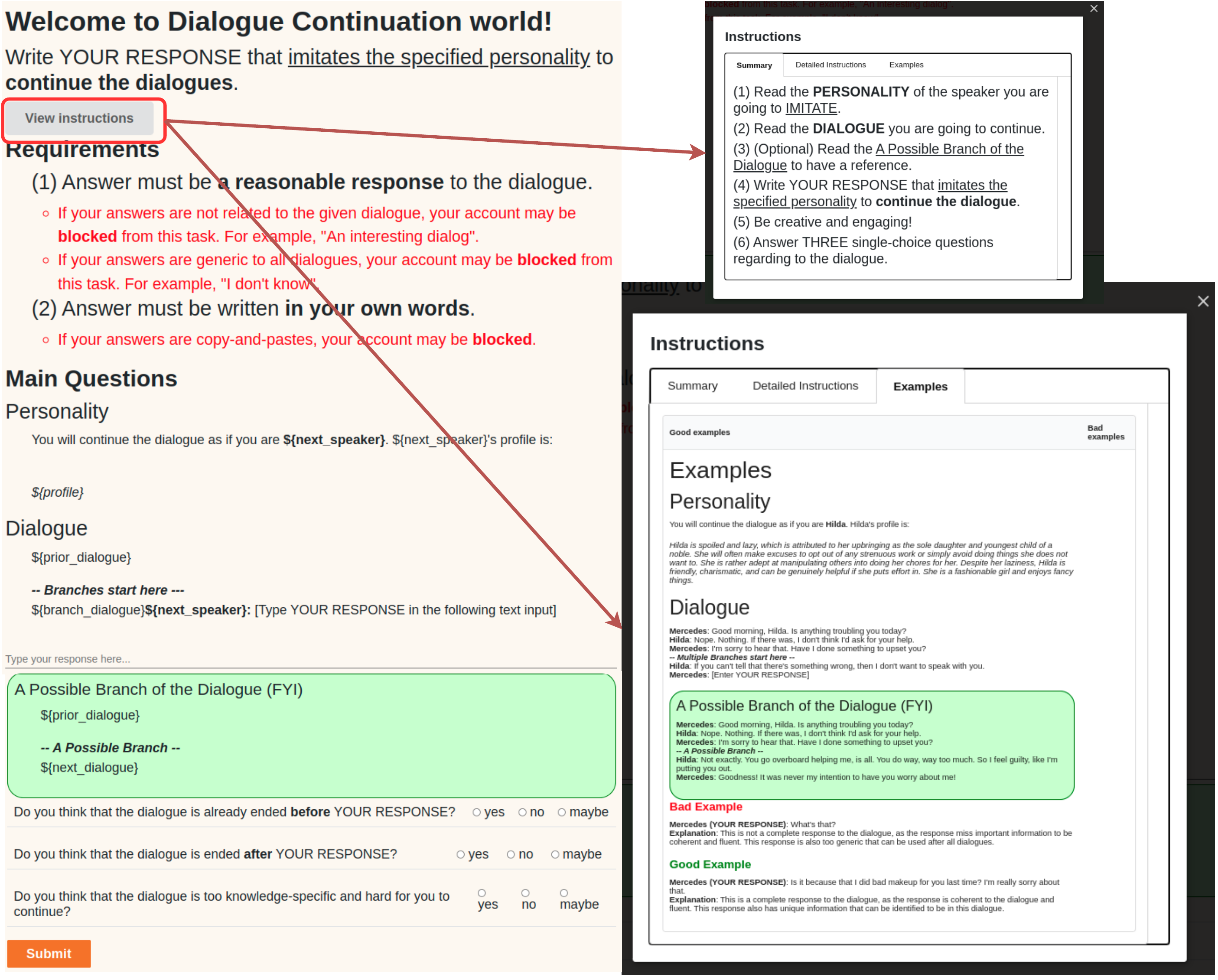}
    \caption{Screenshots of the interface we show the annotators to write responses for the {\bf remaining stages} data expansion. We gave detailed instruction, requirements, and good/bad examples for reference.}
    \label{fig:data_expansion_interface}
\end{figure*}

\paragraph{Interface - Data Expansion.} We design two user interfaces to launch on MTurk for the first stage and the remaining stages separately of the data expansion process.
The interface used for the remaining stages is shown in Figure~\ref{fig:data_expansion_interface}.
We include detailed instructions about the step-by-step works, examples, and requirements to obey.
We put some information into a button to reduce cognitive burden when writing for multiple hits.

\begin{figure*}[t]
    \centering
    \includegraphics[width=.95\linewidth]{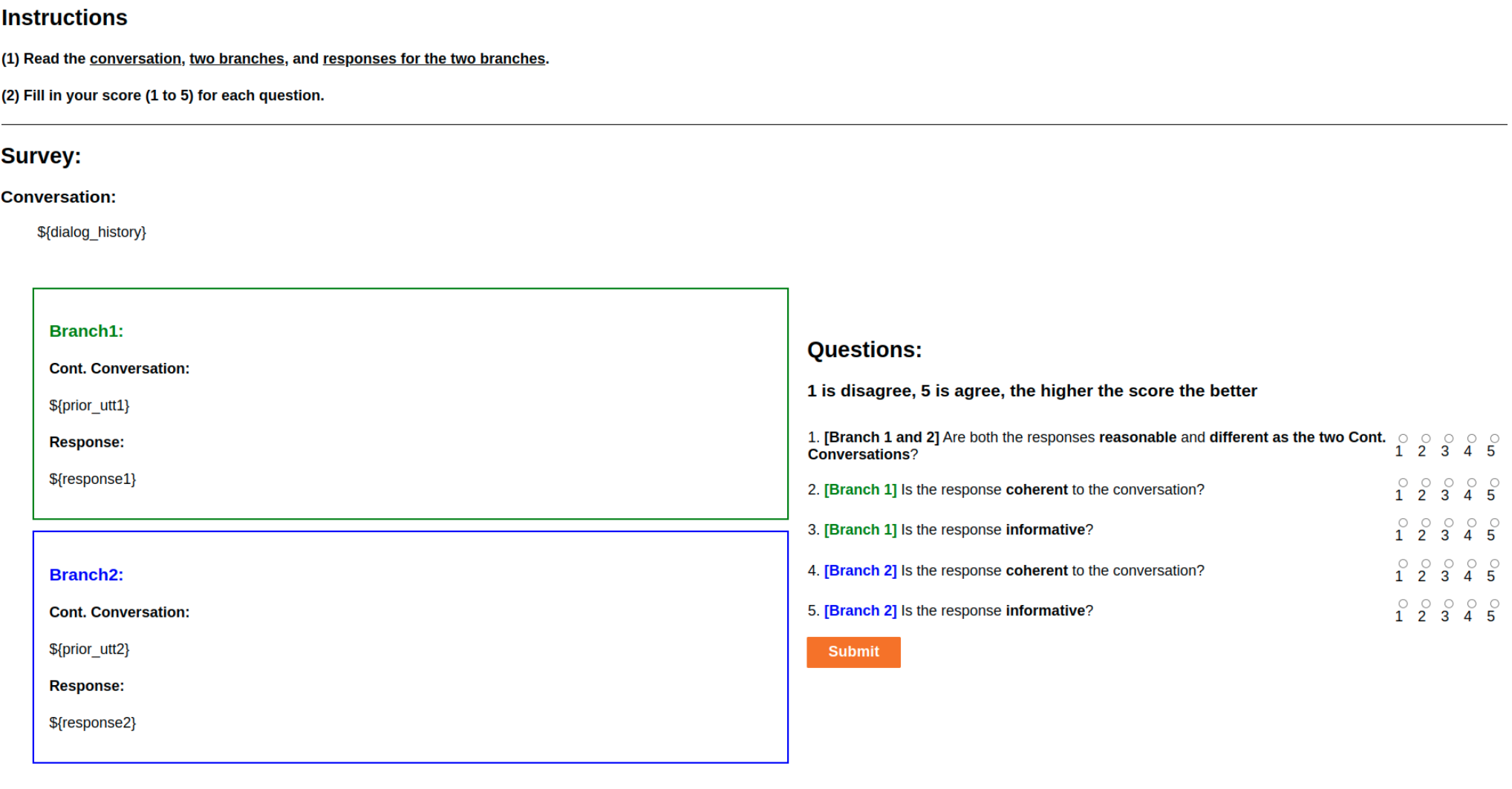}
    \caption{Screenshot of the interface we show the annotators to evaluate generated responses. We gave instruction with five questions each hit.}
    \label{fig:human_eval_interface}
\end{figure*}

\paragraph{Interface - Human Evaluation.}
\label{apx:human_eval_form}
Our used human evaluation form is shown in Figure~\ref{fig:human_eval_interface}.

\paragraph{Setup and Payments.}
We collect the expanded dataset and evaluate generated responses via MTurk, a crowdsourcing platform.
We obtained consent from workers by showing them the study purpose before they agree to do the annotations.
We set additional restrictions of location to United States and Canada.
We pay the annotators from 16-18 US dollars per hour according to the difficulty of the collection stage (remaining stages are more difficult than the first stage).
The payments are made to be higher than the law's minimum wage 15 US dollars per hour in California in 2022 and 15.5 US dollars per hour in 2023, which are the highest among the US states.

\subsection{Evaluation Metrics}
\label{apx:metrics}
Here we discuss more about our selection of evaluation metrics.

\paragraph{Fluency.}
The predicted next utterance should be both coherent to the previous turn and consistent with the dialogue history.
We evaluate the extent of coherence by perplexity and reference-based metric BLEU~\cite{papineni2002bleu}.
For nodes with multiple childs we use multiple references when computing BLEU metrics.
Although that BLEU may not be well correlated with human intuition in conversation~\cite{liu2016not}, we use it for reference as it is still widely used in dialogue generation.
The perplexity (PPL) is considered to be the less the better, whereas BLEU is the higher the better.

\paragraph{Diversity.}
A dialog model can suffer from the generic issue that given different dialogue history and previous turn, the predicted utterance is similar, such as ``I'm sorry''.
We adopt distinct-N scores (Dist1 and Dist2) to evaluate this dimension by considering the percentage of distinct n-grams within the total number of n-grams in the corpus-level~\cite{li2016diversity}.
However, the distinct-N scores are not always the higher the better. We can think about this in a intuitive example, if we randomly sample words from a uniform distribution, the distinct-N score can be high but meaningless.
We anticipate a good distinct-N score is in a similar range as the score evaluated on human written responses.

\subsection{Evaluate Human Responses}
\label{apx:eval_human_responses}
The PPL on human written responses are evaluated by an oracle that will predict a uniform distribution over all human written responses $y$ given the same $(DH,x)$.
The BLEU scores on human written responses are evaluated on data examples with multiple possible responses and the response to be evaluated is hold out from the reference set. Otherwise, the BLEU scores will be 100 since the response to be evaluated is within the reference set.

\subsection{Experiment Details}
\label{apx:exp_setup}
\paragraph{Model architecture.} We use DialoGPT-small with 117M parameters and T5-base with 250M parameters.
DialoGPT model is based on the GPT model architecture (a transformer decoder) but pretrained on conversation-like dataset such as Reddits.
T5 model uses the transformer encoder-decoder architecture and is pretrained on web-extracted text from Common Crawl, which is a publicly-available web archive of scraped HTML files.
The maximum tokens allowed as the input are $256$.

\paragraph{Hyperparameters.} For hyperparameter search, we tried the learning rate from $\{$5e-5,2e-5,1e-5$\}$ and the batch size times gradient accumulation steps from $\{$32,64,128$\}$. We found out that using a learning rate 1e-5 and batch size 64 can generally fine-tuning a model well with different learning algorithms in our experiments.
We train each model with different combinations of setups for single run.

\paragraph{Data preprocessing.} For data preprocessing, we have tried to utilize the original case and punctuation, transform all words into lower case, or meanwhile remove all punctuation.

\paragraph{Computation Resources.} Each model is train on one Titan RTX or one RTX A6000 and costs around five hours.

